\documentclass{article}

\usepackage{graphicx} 
\usepackage[utf8]{inputenc}
\usepackage{amsmath,amssymb}
\usepackage{booktabs} 

\usepackage[round]{natbib}
\usepackage{geometry}
\usepackage{setspace}
\usepackage[hidelinks]{hyperref}

\newcommand{\fSRd}{f(S \mid R,\, d)}

\title{The Differential Meaning of Models: A Framework for Analyzing the Structural Consequences of Semantic Modeling Decisions}

\author{
    Zachary K. Stine\thanks{Corresponding author: \texttt{zstine@uca.edu}} \\
    Department of Computer Science and Engineering \\ 
    University of Central Arkansas \\
    \and 
    James E. Deitrick\\
    Department of Philosophy and Religious Studies \\ 
    University of Central Arkansas
}

\date{August 2025}

\begin{document}

\maketitle

\begin{abstract}
The proliferation of methods for modeling of human meaning-making constitutes a powerful class of instruments for the analysis of complex semiotic systems. However, the field lacks a general theoretical framework for describing these modeling practices across various model types in an apples-to-apples way. In this paper, we propose such a framework grounded in the semiotic theory of C. S. Peirce. We argue that such models measure latent symbol geometries, which can be understood as hypotheses about the complex of semiotic agencies underlying a symbolic dataset. Further, we argue that in contexts where a model's value cannot be straightforwardly captured by proxy measures of performance, models can instead be understood relationally, so that the particular interpretive lens of a model becomes visible through its contrast with other models. This forms the basis of a theory of model semantics in which models, and the modeling decisions that constitute them, are themselves treated as signs. In addition to proposing the framework, we illustrate its empirical use with a few brief examples and consider foundational questions and future directions enabled by the framework.
\end{abstract}


\section{Introduction}
\label{sec:introduction}

The use of computational methods to analyze artifacts of human meaning-making represents a disciplinary collision in which research questions that have traditionally been the domain of the humanities are reinterpreted within a mathematical medium. Such methods primarily take the form of models, which learn a mapping from an input space of symbolic artifacts to a latent space that encodes a distributional semantics inferred from a training set of symbolic artifacts. Such semantic models represent a novel and complex type of scientific instrument, enabling the positing of novel questions at novel scales. \cite{Mallory2025} points out that the creation of novel instruments like the telescope and electron microscope necessitated new theoretical resources to make sense of (a) what the instruments measure and (b) what sorts of claims can be made with those measurements, and the same is required for semantic models. Without such theoretical resources, Mallory argues, it is difficult to make clear claims with the instrument, undermining its scientific utility.  

While semantic models share core properties with various other kinds of instruments, their usage is complicated in cases where they are applied to complex semiotic objects such that verifying the correctness of the instrument's measurements depends on the perspective from which the object is viewed \citep{stine2025semioticcomplexity}. In other words, the measurement of symbolic artifacts with semantic models is subject to something like the observer effect---one cannot ``look" through the instrument and neutrally extract the state of some system. Rather, the instrument must construct, and therefore alter, the state of the system as a byproduct of measuring it. In this light, a description of semantic models must also be what von Foerster calls a ``description of the `describer' or, in other words, ... a theory of the observer" \citep{vonFoerster_epistemology}. For semantic models, measurements are interpretations. Therefore, understanding the measurements we make with them requires understanding the interpretive dispositions that are latent within a model's particular configuration.

In this paper, we lay out a framework in which we can hold up various models side by side, and describe how the views their measurements encode differ from each other. In this sense, our proposed framework can be read as a contribution to what \cite{saxon2024benchmarks} call ``model metrology," with the difference that we are not primarily concerned with evaluation and benchmarks, owing to our emphasis on modeling contexts where what is correct is contingent on an interpretive perspective. Thus, we are after a kind of computational hermeneutics in which we can describe the latent interpretive dispositions that emerge from the vast number of modeling decisions that constitute a semantic model. Our goal is not to theorize the internal mechanics of models, but to theorize the functional, interpretive differences between any set of models. 

In section \ref{sec:theoretical_background}, we lay out a theoretical background and clarify our assumptions about semantic models. In section \ref{sec:semantic_models} we define a formal framework in which we can model the semantics of a semantic model with respect to how it constrains relationships between symbols types. We then build on this framework in section \ref{sec:model_semantics} to construct a methodology for relating the interpretive dispositions of various models. In section \ref{sec:illustrations} we briefly illustrate how the framework may be used to make claims. Finally we consider a few implications of the framework and some important questions to begin answering with it in section \ref{sec:Discussion}.
  
\section{Theoretical background and assumptions}
\label{sec:theoretical_background}

Part of the difficulty in understanding how to approach semantic modeling decisions comes from a lack of clarity around what the modeling task actually is, which is often left undefined or overly vague. This is unfortunate, because instruments are useful insofar as their internal states accurately reflect states of their external environments \citep{Mallory2025}. For example, suppose the air in a room has been in states $s_1$, $s_2$, and $s_3$ at three points in time corresponding to periods during which the air was cold, cool, and hot. If a thermometer, $t$, produced measurements of those states as $t(s_1)$, $t(s_2)$, and $t(s_3)$, then we would reasonably expect that $t(s_1) < t(s_2) < t(s_3)$. We would also expect that $|t(s_1)-t(s_2)| < |t(s_1)-t(s_3)|$ and that $|t(s_1)-t(s_2)| < |t(s_2)-t(s_3)|$. In this example, the instrument is useful insofar as it preserves the geometry of environmental states within the geometry of its internal states. In this section, we argue that we should similarly understand semantic models in terms of the geometries they impose on symbolic state spaces.

By ``semantic model," we mean any model which takes a representation (\textit{i.e.}, signifiers) of a set of objects---symbol types---and maps it to a new representation, constructed from inferred relationships between the variables of the original representation. This definition accommodates a broad class of models, ranging in complexity from simple co-occurrence models to transformer-based neural networks. 

\subsection{The shape is the meaning}

We assume that what is meant by ``semantic" in the context of computational systems is geometry, topology, structure, \textit{etc}. That is, the ``meaning" of a set of objects is the set of relationships between them. In this sense, we can think of the computational meaning of a set of symbols to be the shape of the constellation formed by specifying how near or far each pair of symbols are from each other. Importantly, we assume that this shape does not depend on the particular coordinates and dimensions in which it is instantiated within a model, but can be abstracted into something like a network where vertices correspond to symbols and edges correspond to measures of symbol relationships. For similarity measures, such edge weights are larger when two symbols are nearer, and for dissimilarity measures, they are larger when two symbols are farther apart.

The idea that computational meaning is encoded by geometry is not novel \citep[\textit{e.g.},][]{widdows2004geometry, kozlowski2019geometry_of_culture, STOLTZ2021101567}. It is an implicit assumption of the distributional hypothesis, itself an outgrowth of structuralist thought \citep[for more detailed accounts of this conceptual lineage, see][]{lenci2023distributional, weatherby2025language}. In particular, the distributional hypothesis can be read as an operationalization of the Saussurean notion that signs signify their differences \citep{sahlgren2008distributional}. For Saussure, ``[l]anguage is a system of interdependent terms in which the value of each term results solely from the simultaneous presence of the others," and their ``differences carry signification" \citep{saussure1916}.

It is worth considering why we have an ever-evolving range of models if all they need to do is calculate relationships between symbols. In other words, what is deficient about co-occurrence models that necessitates the use of more complex models, such as artificial neural networks? We think of it like this: Particular datasets, functioning as particular sign representations, are understood to be a sample from a broader underlying sign structure. Semantic models are not typically used to measure a dataset itself (as the surface-level statistics of its variables), but to measure something implied by a dataset. The way each model constructs its picture of this implied sign structure is what defines its interpretive disposition.

\subsection{Symbol structures require ends}
\label{subsec:symbols_and_ends}
We assume that, while symbolic artifacts constrain their interpretation, their meaning does not solely inhere within them, nor does it solely inhere in the intentions of the humans who fashioned those artifacts \citep[see][]{klein2025provocationshumanitiesgenerativeai}. Rather, we assume that meaning arises when symbolic artifacts are encountered by perspectives, and that variation among these perspectives will produce variation in how symbols are structured. This is consistent with the semiotic theory of C. S. Peirce, for whom a sign is ``something which stands to somebody for something in some respect or capacity" \citep[][2.228]{Peirce_1931}. In this framework, perspectives are a necessary aspect of signification, functioning as the ``somebody" that must be specified to resolve what it is a sign may stand for. In the previous section, we argued that for computational models, what signs stand for are their relationships to each other. If a dataset of symbolic artifacts constitutes signs, then semantic models function as a mechanical ``somebody" that determines the geometry of the signs, which is what they are assumed to stand for in such models.

How is it that computational models may usefully act as a perspective despite lacking the complex semiotic agency that humans have? We assume that the way a semiotic agent imbues representations with meaning follows from the values, desires, and goals of that agent---that it is the ends which imbue the means with meaning \citep[see][pp.~16 and 72]{Kohn_2013}. What a sign stands for, then, is resolvable with respect to some well-defined ends, such as solving a coordination problem \citep{fusaroli2012} or, more broadly, maximizing the persistence of a system through time \citep{Kolchinsky2018}. Being machines, computational models differ from humans with respect to ends in that ``[m]achines, as material objects, are means to achieve ends that are, by definition and design, external to them" \citep[][p.~90]{Kohn_2013}. While human ends may also be externally influenced, humans are capable of altering and layering their ends, as evidenced by game-playing, where games are constrained ends \citep[][pp.~4-5]{nguyen2020games}.

If ends are the minimum ingredients necessary for a sign-interpreting perspective, then the objective functions underlying semantic models can be read as mathematically-specified perspectives or the fixed rules for the game. However, the distributional game that semantic models instantiate is a kind of mirror game---given a record of moves (\textit{i.e.}, symbolic artifacts) made by semiotic game-players, the distributional game can be understood as having the goal of inferring the underlying symbol-using strategy implied by the observed moves. In this sense, the sign structures inferred by models are hypotheses about the structure of the semiotic game being played within a dataset. Different ways of specifying the ends of the distributional game within semantic models constitute different ways of constraining these ends, which may produce different interpretive dispositions. In this light, we might think of semantic models as doing a kind of inverse reinforcement learning, in that a model has access to the moves made by humans playing various semiotic games and learns a symbol structure that encodes a hypothesis of what the underlying, distilled game being played is.

This is one way of thinking about the usefulness of semantic models: as an instrument that measures the ways that humans structure their symbol-usage, which, in turn, is an index of the semiotic games they engage in, and we might relate these semiotic games to notions like discourse or even culture. Thus, when large language models generate text from their inferred latent geometries, they convert semiotic strategies into actions, simulating culture by simulating semiotic gameplay \citep[see][]{Mallory2025, Farrell_Gopnik_et_al_2025}.

\subsection{Modeling complex semiotic systems}

Given that human ends are not fixed and are not identical across individuals, it is reasonable to assume that large collections of symbolic artifacts will not have a singular symbol structure, but a complex mixture of various symbol-structuring strategies. However, a semantic model can only encompass a single symbol structure, and so must project onto the mixture of semiotic games, functioning as tokens, a single underlying game, functioning as an abstraction of those tokens to a type, what Kohn calls ``upframing" \citep[][pp.~178-179]{Kohn_2013}. At the same time, human semiotic strategies necessitate being understood by others, such that an individual's symbol-usage reflects not only their own ends, but also their understanding of how other individuals are likely to interpret the symbols they deploy. Hence, the signs that emerge out of semiotic systems are related to complex systems of agencies, and these semiotic agencies are summarized by semantic models according to the particular way in which the distributional game is encoded by their specification. 

For such cases, more information is gained about the semiotic agencies echoed in symbolic artifacts by viewing them through multiple perspectives that are internally reasonable, but which diverge from each other \citep{stine2025semioticcomplexity}. By comparing the sign geometries of different models, one can compare their views. By measuring the relationships between the sign geometries inferred by different models, a change in scope of a semiotic hierarchy is effected in which models (as sign geometries) become points within a higher-order sign geometry. Such a model geometry may be understood as encoding multiple perspectives of the lower-order sign system, thereby widening the semantic aperture of the instrument across an interpretive ecology of models, leading to a widening of the space of claims that can be made with the instrument.

To summarize, our assumptions are as follows: Semantic models interpret the meaning of signs as geometries. Sign geometries depend on the interpretive disposition of a perspective, encoded by ends. The ends of semantic models are to infer a plausible sign geometry from which observed symbolic artifacts emerge. Therefore, the sign geometries semantic models infer are constrained by the symbolic artifacts they are trained on \textit{and} by a model's specification. A model is a collection of modeling decisions. The goal of the framework we construct in the following sections is to give us a precise language for asking how sign geometries change as a result of changing modeling decisions.

\section{A theory of semantic models}
\label{sec:semantic_models}

With this reasoning and motivation laid out, we can now state our assumptions about semantic models with some more precision. We organize our theory around two kinds of functions, or maps: representation maps, which send input representations to latent representations, and the structural map, which sends any representation to a geometry of symbol types that is abstracted from the particular coordinates and dimensionality of a representation. Representation maps send representations to spaces in which semantic structure is expressed across at least two kinds of objects, while the structural map sends representations to a space in which this structure is simplified to a single set of objects, which bears some resemblance to metric spaces. Our primary goal in introducing the structural map is to define a common medium in which the semantics underlying different representations can be compared directly, rather than by proxy via performance measures.

\subsection{The structural map}

Let $S$ be a set of $n$ symbol types, $\{s_1, \ldots, s_n\}$. Let $R$ be a collection of symbolic artifacts (e.g., sequences of text) in which the elements of $S$ are instantiated, with $R[S] = \{R[s_1], \dots, R[s_n]\}$. According to the assumptions laid out in the previous section, the target variable that a semantic model predicts is the set of relationships among the elements of $S$ that are hypothesized to underlie the semiotic agencies that produced $R$. That is, the target variable is the differential meaning of $S$---its geometric structure---given how its elements are instantiated within a representation $R$ and a relation measure $d$, defined as 
\begin{equation}
f(S \mid R, \, d) = \big\{\, d\!\left(R[s_i], \, R[s_j]\right) \; : \;  s_i, s_j \in S, \, i \neq j \, \big\}.
\end{equation}
 We use $d$ to highlight the connection to differential meaning, but $d$ could just as well be a similarity measure. In either case, we can think of $f(S \mid R, \,d)$ as defining edge weights in a graph with vertices given by $S$, forming a semantic network whose topology is the meaning of $S$ according to $R$ and $d$. 
 
We refer to $f$ as the \textit{structural map}, which maps a particular combination of modeling decisions, $(S, R, d)$, from the space of modeling decisions, $\mathcal{S} \times \mathcal{R} \times \mathcal{D}$, to a particular structure, $G$, within the space of structures, $\mathcal{G}$. A semantic model thus requires a minimum of three choices: a particular set of symbol types $S$ from the space of all sets of symbol types $\mathcal{S}$, a particular representation $R$ from the space of all representations $\mathcal{R}$, and a particular relation measure $d$ from the space of all relation measures $\mathcal{D}$. However, the choice of $S \in \mathcal{S}$ constrains the representation space $\mathcal{R}$ to only those representations which correspond to $S$, denoted by $\mathcal{R}_S$, and similarly constrains the structural space $\mathcal{G}$ to only structures on $S$, denoted by $\mathcal{G}_S$. For this reason, $S$ can be seen as a modeling decision that precedes any other and so indexes a particular $f_S$ from the structural map as a family of functions, $\{f_s : \mathcal{R}_S \times \mathcal{D} \to \mathcal{G}_S\}_{S \in \mathcal{S}}$. Here, $S$ is treated as a fixed parameter rather than a free input that can vary within a particular mapping.

Importantly, all structures on $S$ are relative to both $R$ and $d$. That is, we do not neutrally look through particular representations and relation measures to see the semantics of $S$, but rather see the semantics of $S$ as ``interpreted" by $R$ and $d$.  If signs stand for something to somebody, then $R$ can be understood to stand for $\fSRd$ to $d$. Here, $d$ is necessary to collapse the various potential structures on $S$ that are implicit in $R$ onto a single structure acting as a particular, fixed interpretation. As an analogy, we can think of representations like camera film that has been exposed and relation measures like the process of developing exposed film into fixed photographs. Representations, like the camera film, contain information that may be developed into related, but different pictures when developed, depending on the developing choices a photographer makes, corresponding to $d$. We may have various kinds of cameras that use different sizes of film, but we can develop onto a fixed print size to make comparisons between them.

While the elements of $S$ must belong to the same level within a semiotic hierarchy, representations are assumed to consist of at least two sets of objects that belong to different levels of the hierarchy. 

To make this clear, we will illustrate with one of the simplest kinds of semantic models: word counts matrices. Suppose that $R_a$ is a matrix of word frequencies arranged into $m$ rows, representing documents, and $n$ columns representing word types, which constitute two sets of objects at different levels of scope. Let $d_{\cos}$ be the cosine distance. There are two possible sign systems that we might model by calculating the cosine distance between symbol representations: the objects corresponding to the $m$ rows or those corresponding to the $n$ columns. In other words, $R_a[S]$ is not inherent to a representation, but requires a choice to be made about what compositional layer of the representation will serve as the symbol types. 

If we think of symbol types as the observations of a dataset, which are described by a set of attributes, then $S$ could either be the documents or the word types. Both are abstractions that describe each other on the basis of token frequencies. If the observations are taken to be the documents, then each observation is described by the extent of a word type within it. On the other hand, if the word types are taken to be the observations, then each observation is described by the extent of a document within it.  

If $S$ is chosen to correspond to the $m$ documents, and $d$ is chosen to be the cosine distance, then computing the cosine distance between each pair of $n$-dimensional document vectors amounts to computing $f(S \mid R_a, d_{\cos})$---the structure or interpretation of $S$ relative to $R_a$ and $d_{\cos}$. Keeping $S$ fixed, but varying the representation or relation measure would presumably result in different structures, reflecting the interpretive differences between them. If varying $R$ or $d$ produces identical structures on $S$, then those variations can be said to be structurally equivalent relative to $S$. For example, if $f(S \mid R_a, d_{\cos})$ has the equivalent structure given by $f(S \mid R_b, d_{\cos})$, then $R_a$ and $R_b$ are semantically (\textit{i.e.}, structurally) equivalent relative to $S$ and $d_{\cos}$.   

\subsection{Representation maps}

Thus far we have considered the choice of representation $R$ to be the result of a single modeling decision. However, representations themselves may be the product of many modeling decisions such that the single choice of $R$ is better understood as the set of all choices which produced $R$. In this sense, representations may be understood as existing within a representational chain in which each representation extends the previous one through some additional modeling decisions. In the example above, choosing the document-term matrix $R_a$ means choosing all of the modeling decisions necessary to create $R_a$, extending back to the choice of some initial digital representation to use as data. To make it clear that a representation is the product of such modeling decisions, we now introduce the idea of \textit{representation maps}. 

Let $\theta$ be a representation map, $\theta: \mathcal{R} \to \mathcal{R} $, that takes an initial representation $R \in \mathcal{R}$ as input and sends it to another representation $\theta R \in \mathcal{R}$. Representation maps can be thought of as arbitrarily long lists of sequential instructions, where each instruction constitutes a modeling decision (regardless of whether or not the modeler is aware of them). 

What we typically think of as semantic models are representation maps where their output representations are understood to have changed the semantics from the input representation in useful ways. We refer to them in this way to make it clear that, in this framework, semantic models are not understood to take non-semantic representations and imbue them with semantics---since every representation constrains a space of implicit semantics---but rather act primarily to change the representation and so change the semantics as a result. A semantic model may map representations with surface-level semantics to latent semantics, but it does not confer semantics upon representations nor increase the amount of semantics in any ordinal sense.

Going back to the example, the choice of the document-term matrix $R_a$ is really the choice of an initial representation $R$ and all of the choices for how to turn $R$ into $R_a$, given by the representation map $\theta_a$ such that $R_a = \theta_a R$. In this case, $R$ might be a text file containing the $m$ documents as strings of tokens*\footnote{Even this text file constrains a range of possible structures that can be collapsed onto a single structure by the choice of a relation measure like the edit distance between strings}. This notation is meant to emphasize that the choice of $R_a$ is decomposable into the choices of both $\theta_a$ and $R$. Additionally, $\theta_a$ may itself be decomposed into intermediary representation maps such as case normalization, $\theta_{a1}$, tokenization, $\theta_{a2}$, stopword removal, $\theta_{a3}$, and the counting of word frequencies and their formulation into a matrix, $\theta_{a4}$, so that $\theta_a R$ can be expanded to $\theta_{a4}(\theta_{a3}(\theta_{a2} (\theta_{a1} R)))$. 

A slightly more complex representation map, $\theta_b$, can be made by extending $\theta_a$ with the following modeling decisions: Choose an integer $k$ such that $k < n$, and then convert the $m\times n$ document-term matrix to an $m \times k$ matrix by singular value decomposition. The resulting representation $R_b = \theta_b R$ consists of the $m$ documents from the initial representation and of $k$ latent variables. These latent variables take the form of linear combinations of the word types from $R_a$ (which are expressed in a related representation as a $k\times n$ matrix of weights). Just as $\theta_a R$ consists of two sets of objects which could function as the set of symbol types, so too does $\theta_b R$ where either the $m$ documents or the $k$ latent variables could function as $S$. Our goal is to compare the structural consequences of choosing $\theta_a$ or $\theta_b$, and for such a comparison to be possible, the choice of $S$ must be the same for both $\theta_a R$ and $\theta_b R$.

We can now expand the structural map to include the choice of initial representation $R$ and representation map $\theta$ so that $f_S$ maps a point given by $(R, \theta, d)$ to a structure on $S$:
\begin{equation}
    f(S \mid \theta R, d) = \big\{d(\theta R[s_i], \theta R[s_j]     
    \; : \;
    s_i, s_j \in S
    \big\}.
\end{equation}
 
\subsection{Codes}

While representation maps modify representations, it is also possible to think of representations as modifying representation maps. This is the case when a representation map includes parameters that are adjusted in response to a representation through model training. The resulting representation map can then be applied to other representations, structuring them according to the training representation. Rather than draw out latent structures in a representation, we can understand this as a representation map drawing out a latent code from the training representation.

For example, let $R_\alpha$ be a representation on which a model is trained such that the representation map $R_\alpha  \theta$ produces representations that depend on model weights within $\theta$ that have been adjusted in response to $R_\alpha$. The resulting map $R_\alpha \theta$ can be applied to a different representation $R_\beta$ to create a new representation $(R_\alpha \theta)R_\beta$, which is created on the basis of the attribute semantics implicit in $R_\alpha$ according to the specification of $\theta$. In this way, the representation map given by $R_\alpha \theta$ enables us to ask how $R_\beta$ may be structured according to a code that emerges from the combination of $R_\alpha$ and $\theta$. That is, one can ``read" $R_\beta$ through a lens that is shaped by the combination of representation $R_\alpha$ with the interpretive disposition of $\theta$. 

If $f(S \mid (R_\alpha\theta)R_\beta, d)$ is structurally equivalent to $f(S \mid \theta R_\beta, d)$---where $\theta R_\beta$ is notational shorthand for $(R_\beta \theta)R_\beta$---then $R_\alpha$ and $R_\beta$ are said to have equivalent codes relative to $R_\beta[S]$, $\theta$, and $d$. 

\section{A theory of model semantics}
\label{sec:model_semantics}

In this framework, understanding the semantic consequences of modeling decisions does not require understanding the computational machinery behind them, which can be highly complex, as in large language models. Instead, we focus on comparing modeling decisions relative to alternatives by examining the structural differences that arise across them. This allows us to make clear, formal claims about semantic models by describing how variations in decisions produce variations in structure, rather than by analyzing the internal mechanics of the models themselves. So while it is often useful to work with a latent representation $\theta R$ in many research contexts, mapping such a representation to an explicit structure permits a functional understanding of $\theta$ in terms of how the structure changes with respect to changes in $\theta$. 

We can think of $f(S \mid \theta R, d)$ as a perspective-dependent sample of an idealized $f(S \mid R)$---the meaning of $S$ according only to its representation within symbolic artifacts, and therefore, according to the semiotic agencies that produced them, independent of representation maps and relation measures. While such a complete description may be impossible, it is useful to conceptualize it as the total information about $R[S]$ that emerges when viewing it from all plausible perspectives:
\begin{equation}
    f(S \mid R) \approx \{f(S \mid \theta R, d) \; : \; \theta \in \Theta, \, d \in D \},
\end{equation}
as $\Theta$ and $D$ grow to encompass as many ``perspectives" as possible. However, even if it were possible to compute this for all plausible representation maps and relation measures, one would still have to make sense of each individual structure by relating it to the others, and this necessitates another modeling decision: the choice of structural relation measure $\delta$, which indicates the dissimilarity or similarity between two structures. Such a measure could take the form of the Procrustes distance or a measure of correlation \citep[see][for other possibilities]{klabunde2023similarityneuralnetworkmodels}. 

While using an ecology of models across the space of $\Theta$ to view a particular $R$ is useful for modeling the complex system of semiotic agencies that produced $R$, one can also turn this around by using an ecology of representations across the space of $\mathcal{R}$ to better understand the interpretive disposition of a particular $\theta$, which is what we mean by ``computational hermeneutics" \citep[cf.~][]{Mohr2015compherm}. 

More concretely, let $\Sigma$ describe a set of alternatives for the same subspace of modeling decisions, which function here as higher-order symbol types. We are concerned with describing the structural consequences of each element in $\Sigma$ relative to the others against a backdrop of modeling decisions which remain fixed for all structures, which we denote by $\Phi$, and structural relation measure $\delta$. For example, $\Sigma$ could be different choices of $k$, with $\Phi$ comprising all fixed modeling decisions: the choice of initial representation, preprocessing scheme, computing the singular value decomposition, and calculating the pairwise cosine similarity between documents. 

Let $\Phi[\Sigma]$ be the set of resulting sign structures that share the modeling decisions of $\Phi$ in common and differ only according to the elements of $\Sigma$. Then the semantics of $\Sigma$ relative to $\Phi$ and $\delta$ is 
\begin{equation}
f(\Sigma \mid \Phi, \, \delta) = 
\big\{ 
    \delta \! \left(  \Phi[\sigma_i], \Phi[\sigma_j]  \right)   
    \; : \;
    \sigma_i, \sigma_j \in \Sigma
\big\} .
\end{equation}

The elements of $\Sigma$ may correspond to representations, representation maps, relation measures, or even structural relation measures. If the elements of $\Sigma$ yield identical structures, they are structurally inconsequential and can be considered ``meaningless” in this framework. This opens the door to doing a kind of instrument calibration in that it allows the researcher to focus the lens of the instrument such that the sign structures under analysis emerge primarily from a desired set of modeling decisions and not from decisions that are seen as unimportant to the analysis. For example, a researcher may want to only consider models in which the choice of random seed is not structurally consequential, thereby focusing this modeling choice out of the structural picture.

Essentially, we are applying this theory of semantic models to itself, treating models of signs as signs themselves \citep[see][]{ciula2017modelling}. If we think of $f(S \mid \theta R, d)$ as a graph with vertices given by $S$, we can similarly think of $f(\Sigma \mid \Phi, \delta)$ as a graph in which each vertex $\sigma \in \Sigma$ is itself a graph with vertices $S$.

\subsection{The semantics of representation maps}

If a researcher is primarily interested in the semiotic agencies that produced some representation $R$, then they will likely be interested in understanding the extent to which any latent representation $\theta R$ is an index of the interpretive disposition of $\theta$ rather than that of $R$. To make the interpretive disposition of $\theta$ visible, we can view it against a backdrop of alternative semantic models, $\Theta$. 

Let $R[S]$ and $d$ be fixed, corresponding to $\Phi$. Let $\Theta = \{\theta_1, \ldots, \theta_n\}$ be a set of representation maps which vary, corresponding to $\Sigma$. For example, $\Theta$ might correspond to latent semantic analysis, latent Dirichlet allocation, and other semantic models. They may also vary in terms of their hyperparameters or other configurations. The semantics of the representation maps $\Theta$ relative to $\Phi$---$R[S]$ and $d$---and structural relation measure $\delta$ is given by
\begin{equation}
f(\Theta \mid  \Phi, \, \delta) = 
\big\{  
    \delta( \Phi[\theta_i], \Phi[\theta_j]) 
    \; : \;
    \theta_i, \theta_j \in \Theta
\big\},
\end{equation}
where $\Phi[\theta_i] = f(S \mid \theta_iR, \, d)$.

To illustrate, let us return to the example of the $m \times n$ document-term matrix $R_a =\theta_aR$, and the $m \times k$ latent representation $R_b = \theta_bR$. Since $\theta_b$ consists of identical modeling decisions as $\theta_a$ up to a point, any structural difference between $f(S \mid \theta_aR, d)$ and \mbox{$f(S \mid \theta_b R, d)$} must be due solely to the additional modeling decisions in $\theta_b$. That is, $\delta(\Phi[\theta_a], \Phi[\theta_b])$ is an index of the difference between $\theta_a$ and $\theta_b$ in the context of $R[S]$ and $d$ according to $\delta$. 

This gives us a way of calculating the dissimilarity, distance, or similarity between models. This is useful because it allows us to make claims about the effective differences between models without comparing their particular mechanics, which may be exceedingly complex. 

\subsection{The semantics of representations}

While the observations and attributes that constitute a representation $R$ may often function as the sign system under analysis, it also possible to treat representations as signs themselves which can be related to each other within the context of a shared representation map. 

Let $\mathcal{R} = \{R_1, \ldots, R_n\}$ be different representations of $S$, corresponding to $\Sigma$. Let $S$, $d$, and $\theta$ be fixed and so correspond to $\Phi$. Then the semantics of $\mathcal{R}$ relative to $\Phi$ and $\delta$ is given by
\begin{equation}
    f( \mathcal{R} \mid \Phi, \delta) = \big\{
        \delta(\Phi[R_i], \Phi[R_j])
        \; : \;
        R_i, R_j \in \mathcal{R}
    \big\}
\end{equation}
where $\Phi[R_i] = f(S \mid \theta R_i, d)$.

\subsection{The semantics of relation measures}

While much of our interest may lie in characterizing the structural consequences of representation maps and representations, it is also possible to describe the influence of a particular choice of relation measure $d$ on the resulting structure of $S$. Rather than characterize a relation measure in terms of our intuitions about its appropriateness in a given context or with respect to performance measures, we can analyze the extent to which varying $d$ varies the structure of $S$ within the structural map. 

Let $D = \{d_1, ..., d_n\}$ be a set of relation measures that can be applied to the pairs of elements in $R[S]$.  Here, the set of measures $D$ plays the part of $\Sigma$ whereas the choices of $R$ and $S$ correspond to $\Phi$. Then the semantics of $D$ relative to $\Phi$ and a structural relation measure $\delta$ is given by 
\begin{equation}
f\left(D \mid \Phi, \, \delta \right) = 
\big\{ 
    \delta \! \left( \Phi[d_i], \Phi[d_j]\right)   
    \; : \; 
    d_i, d_j \in D   
\big\}.
\end{equation}

where $\Phi[d_i] = f(S \mid R, d_i)$.

\subsection{The stability of model semantics}

We now return to the case where one has a variety of representation maps $\Theta$ which may differ in their internal machinery as well as training data. As we have seen, the semantics of $\Theta$ relative to $\Phi$, consisting of $R$ and $d$, according to $\delta$ is given by $f(\Theta \mid \Phi, \delta)$. Let us consider two extreme possibilities for such an analysis.

First, consider what it would mean if all representation maps in $\Theta$ produced identical structures on $S$. For example, if $\delta$ is the Procrustes distance, then for all $\theta_i, \theta_j \in \Theta$, $\delta(\Phi[\theta_i], \Phi[\theta_j]) = 0$. Unlikely as this might be, interpreting such a case is helpful for understanding the sorts of claims we can make with the framework. Such a case would imply that the choice of $\theta \in \Theta$ has no structural consequences---varying the choice produces no structural variation---so that choosing from $\Theta$ is ``meaningless" according to $R$, $d$, and $\delta$. This would simplify an analysis of $R$ by permitting a researcher to simply work with a single representation map. 

For the second extreme, consider what it would mean if each $\theta \in \Theta$ produced a highly distinct structure of $S$. In this case, one would be confronted by the interpretative pluralism of $\Theta$ and this would complicate the analysis of $R$, expanding it across the various structures. One might begin analyzing how those structures differ, perhaps by computing which elements of $f(S \mid \theta_i R, d)$ diverge most significantly from those of $f(S \mid \theta_j R, d)$.

The most likely case is that some elements of $\Theta$ will lead to similar structures of $S$ while others will differ significantly. In this case, one might determine which $\theta$ is representative of a structural consensus and decide to focus on that representation map. On the other hand, one might decide to remove any $\theta$ that is structurally similar to any other, reducing $\Theta$ down to those which are structurally distinct. 

From these considerations, a foundational question emerges when one considers repeating such an analysis of $\Theta$ for different representations: Are the semantics of $\Theta$ fixed across representations, or are they particular to each representation? For example, let $f(\Theta \mid \Phi_i, \delta)$ be the structure of $\Theta$ relative to representation $R_i$ and relation measure $d$. What would it mean if this structure differed in any significant way from $f(\Theta \mid \Phi_j, \delta)$, the structure of $\Theta$ relative to $R_j$ and $d$? 

If the relationship between semantic models in $\Theta$ is consistent across representations, then the semantics of $\Theta$ are likely to be fixed. Lessons learned about models in one representational context are likely to generalize to others. However, if the relationships between the semantic models in $\Theta$ may change significantly across representational contexts, then the models' semantics are contextual, and their particular geometry relative to a representation is an index of the representation. On the one hand, this would complicate how we make sense of semantic modeling decisions. On the other hand, a structure of $\Theta$ relative to representation $R_i$,  would function as a higher-order latent representation $\Theta R_i$, semantically richer than any individual $\theta R_i$. 

\subsection{Navigating the semiotic hierarchy}

Recall that each state in the symbolic state space of $f(\Theta \mid \Phi, \delta)$ corresponds to a nested structure in which the symbolic state space is defined by $S$ at the ground level, and is defined by $\Theta$ at the next level up. If one then varies representations across $\mathcal{R}$, then those representations define the symbolic state space at the next outer level. Each vertex in such a graph of $\mathcal{R}$ is itself a graph of $\Theta$, which in turn has as vertices graphs of $S$. 

Due to this fractal-like quality, navigating across this semiotic hierarchy can be confusing, since what may function as a symbol at one level may have a different function at another. But we can make the scope of analysis explicit by our definition of the inputs to the structural map. In this example, the semantics of $\mathcal{R}$ at the outermost layer is given by

\begin{equation}
    f(\mathcal{R} \mid \Phi', \delta') = \big\{ 
    \delta'(\Phi'[R_i], \Phi'[R_j])
    \; : \; R_i, R_j \in  \mathcal{R} \big\},
\end{equation}
where $\Phi'[R_i] = f(\Theta \mid \Phi, \delta)$ in which the semantics of $\Theta$ at the next layer down is given by
\begin{equation}
    f(\Theta \mid \Phi, \delta) = \big\{
    \delta(\Phi[\theta_k], \Phi[\theta_\ell])
    \; : \;
    \theta_k, \theta_\ell \in \Theta
    \big\},
\end{equation}
where $\Phi[\theta_k] = f(S \mid \theta_kR_i, d)$.

Such an analysis spans three levels in the hierarchy with $\Sigma$ corresponding to $S$ at the first level, $\Theta$ at the second, and $\mathcal{R}$ at the third. Note that $\delta$ and $\delta'$ may be the same structural relation measure, and the difference in notation is meant to differentiate the hierarchical level.  

\subsection{Meaninglessness as structural reference points}

The structural difference between two modeling decisions becomes informative in the context of  alternative modeling decisions.  The information conveyed by $\delta(\Phi[\sigma_i], \Phi[\sigma_j])$, for example, is minimal when $\Sigma$ only consists of $\sigma_i$ and $\sigma_j$ such that $f(\Sigma \mid \Phi, \delta)$ consists of a single value if $\delta$ is symmetric. The more elements that $\Sigma$ comprises, the more informative $\delta(\Phi[\sigma_i], \Phi[\sigma_j])$ becomes, as the background against which it is interpreted grows. The relationships between signs may thus function as signs as well. 

Two reference points across structural spaces may help provide additional context: \textit{random models}, in which all values in $f(\Sigma \mid \Phi, \delta)$ are randomly drawn from a uniform distribution, and \textit{null models}, where all values of $f(\Sigma \mid \Phi, \delta)$ are identical. Both kinds of models are meaningless in the sense that they are not the products of any actual distributional information within symbolic artifacts. Random models are meaningless in this sense despite possessing sign differentiation, while null models eliminate differences altogether. 

These two extremes can be seen as information bounds in a structural space by considering the uncertainty in drawing the values of $f(\Sigma \mid \Phi, \delta)$, which is maximal for the random models and minimal for the null model.

Note that, while there can be many random models for a structural space, all null models will have identical shape, since they can only differ in scaling, and so can be thought of as a single model or equivalence class of models.

\section{Illustrations of the framework}
\label{sec:illustrations}

We now illustrate the framework by putting it to use in making some simple but clear claims about a collection of English-language album reviews from the online music publication, Pitchfork, and the probabilistic topic model, latent Dirichlet allocation (LDA) \citep{blei2003latent}. While there is room for human interpretation of the analyses we conduct, our goal here is only to demonstrate the sorts of empirical claims we can make prior to any interpretation of them (which will be the focus of a future article). To this end, our choice of dataset and model type can be considered arbitrary, and these analyses are in no way complete, but serve as methodological sketches of how this framework might begin to be applied as part of larger analyses. 

Our initial representation $R$ takes the form of 24,378 English text sequences, with each sequence corresponding to an album review. Of course, this representation did not spontaneously come into being, but is itself the result of non-computational representation maps within a semiotic chain. Prior to being text, each review may have been represented by the mechanics of keys being typed on, and in turn, those physical keyboard states can be understood as representations of the reviewer's thoughts in response to an album. In this illustration we consider the semantics of the reviews (rather than the word types they consist of), which will constitute $S$. For computational convenience, we let $S$ be a subset of 1,000 reviews from the total collection.

All representation maps that we consider are identical with the exception of two modeling decisions: the number of latent variables (i.e., topics) $k$, and the random seed from which a model's parameters are initialized, $\psi$. We train ten models for each value of $k$, for which we use the following nine values: 5, 15, 30, 50, 100, 200, 500, 750, and 1,000. Each model's random seed was chosen at random. Thus our population of representation maps $\Theta$ consists of 90 elements which have unique random seeds, and which may be grouped into nine sets by $k$. We denote a particular representation map with the choice of $k$ and $\psi$ as $\theta_{k,\, \psi}$. Then the full set of representation maps can be defined by 
\begin{equation}
    \Theta = \big\{ \theta_{k, \, \psi}  
    \; : \;
    k \in K, \, \psi \in \Psi \big\},
\end{equation}
where $K$ is the set of nine values for the latent dimensionality and $\Psi$ is the set of all random seeds. 

Each model is trained on the full collection of 24,378 reviews and then used to represent the 1,000 reviews in $S$ as $\theta_{k, \psi}R[S]$---a set of 1,000 $k$-dimensional probability distributions. Lastly, we use the Jensen-Shannon divergence (JSD) for the relation measure $d$ and the Procrustes distance as the structural relation measure $\delta$.\footnote{Our code is available at \url{ https://github.com/zacharykstine/differential_meaning_of_models}.}

\subsection{The stability of meaning in LDA}

First, we investigate the structural stability of LDA when the random seed $\psi$ is varied. We investigate this within each subset of models that share the same value of $k$. Our goal is to answer the question: Within each of the nine values of $k$ we investigate, how sensitive are the semantics of $S$ on the choice of the random seed? We might reasonably like for the choice of random seed to be meaningless in the sense that we do not want the resulting semantics from an LDA model to vary wildly with varying the random seed. A similar analysis of stability in topic models is given by \cite{greene2014_stability}, which focuses on the stability of latent attributes (\textit{i.e.}, topics). Here, the particular form that latent attributes may take is considered irrelevant. That is, it is possible for two topic models to appear distinct on the basis of their topics while being functionally identical in how they structure documents.

We start with the subset of $\Theta$ that all share the modeling decision $k = 5$, which we will denote by $\Theta_{k=5}$. This subset of models differs only in the choice of random seed, $\psi$. We then compute $f(S \mid \theta_{k, \psi_i} R, \, d)$ for each $\theta \in \Theta_{k=5}$, resulting in ten structures on $S$, each given by a $|S|\times |S|$ matrix of JS divergences. While it would be reasonable to only calculate the upper triangle of this matrix (given that the JSD is symmetric) we use the full matrix since one might wish to compare such a structuring to one in which an asymmetric measure was used \citep[see][on the usefulness of asymmetric relation measures]{Chang2020_divergence}. 

In order to determine how much these ten structures differ from each other (which can only be a consequence of the differing random seeds), we then compute $f(\Sigma \mid \Phi, \delta)$ where $\Sigma$ corresponds to the random seeds of $\Theta_{k=5}$; $\Phi$ corresponds to $S$, $R$, $d$, and $k$; and $\delta$ is the Procrustes distance. We could write this as $f(\Theta_{k=5} \mid \Phi, \delta)$, but since $k$ is fixed among the elements of $\Theta_{k=5}$, we can more clearly write this as $f(\Psi_{k=5} \mid \Phi, \delta)$ where $\Psi_{k=5}$ denotes the random seeds of the models in $\Theta_{k=5}$. This emphasizes the interpretation of $f(\Psi_{k=5} \mid \Phi, \delta)$ as the semantics of the random seeds according to $\Phi$ and $\delta$. That is, 
\begin{equation}
    f(\Psi_{k=5} \mid \Phi, \delta) = \big\{\delta(\Phi[\psi_i], \Phi[\psi_j])
    \;:\;
    \psi_i, \psi_j \in \Psi_{k=5}
    \big\}
\end{equation}
where $\Phi[\psi_i] = f(S \mid \theta_{k=5, \psi_i}R, d)$.  Since there are ten elements in $\Psi_{k=5}$, its structuring consists of 45 values (since the Procrustes distance is symmetric and we do not calculate self-similarity). 

\begin{figure}
  \centering
  \includegraphics[width=\linewidth]{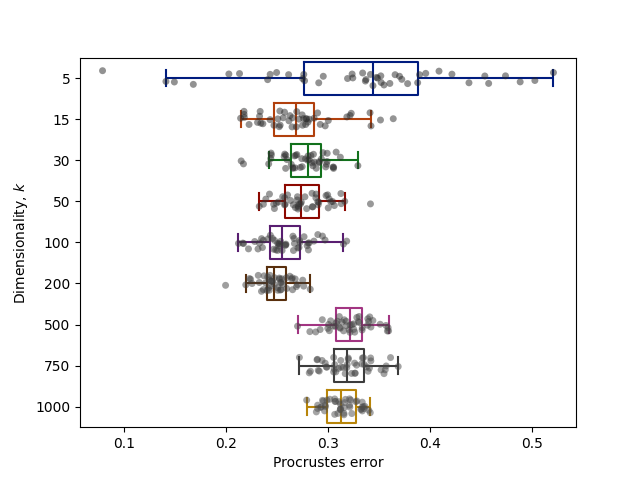}
  \caption{The pairwise Procrustes distance distributions calculated between the dissimilarity matrices of models with the same dimensionality, $K$.}
  \label{fig:procrustes_boxplot_for_each_k}
\end{figure}

We then repeat this process for each of the other values of $k$, computing $f(\Psi_{k=15} \mid \Phi, \delta), f(\Psi_{k=30} \mid \Phi, \delta)$, and so on. The results can be seen in \ref{fig:procrustes_boxplot_for_each_k}. The mean and standard deviation for the Procrustes distances for each choice of $k$ are given in \ref{tab:procrustes}. Additionally, we include the same statistics for the Procrustes distance between each model and a set of random models (in which $S$ is structured by randomly chosen values) and a null model (in which the relations between all pairs of $S$ are equal).

\begin{table}
  \caption{Procrustes error between model pairs with the same dimensionality, K.}
  \label{tab:procrustes}
  \begin{center}
  \begin{tabular}{cllllll}
    \toprule
       & \multicolumn{2}{c}{LDA $\leftrightarrow{}$LDA} & \multicolumn{2}{c}{LDA $\leftrightarrow{}$Random} & \multicolumn{2}{c}{LDA $\leftrightarrow{}$Null} \\
    $K$  & Mean & Std dev & Mean & Std dev & Mean & Std dev\\
    \cmidrule(lr){1-1} \cmidrule(lr){2-3} \cmidrule(lr){4-5} \cmidrule(lr){6-7}
     5 & 0.333 & 0.0964 & 0.996 & 0.0013 & 0.995 & 0.0001\\  
    15 & 0.271 & 0.0366 & 0.994 & 0.0009 & 0.992 & 0.001\\ 
    30 & 0.277 & 0.0237 & 0.992 & 0.001 & 0.989 & 0.0009\\ 
    50 & 0.277 & 0.0239 & 0.991 & 0.001 & 0.987 & 0.0011\\
    100 & 0.259 & 0.0244 & 0.988 & 0.0013 & 0.982 & 0.0015\\
    200 & 0.249 & 0.0162 & 0.982 & 0.0012 & 0.972 & 0.0019\\
    500 & 0.321 & 0.0206 & 0.96 & 0.002 & 0.934 & 0.0032\\
    750 & 0.32 & 0.0232 & 0.944 & 0.0015 & 0.907 & 0.0022\\
    1000 & 0.314 & 0.016 & 0.923 & 0.003 & 0.88 & 0.0038\\
  \bottomrule
\end{tabular}
\end{center}
\end{table}

What these results show is that the semantic stability of LDA can be sensitive to the choice of random seed, and of the values of $k$ explored here, the random seed choice is most consequential in the case of $k=5$ and is the least consequential for $k=200$. In other words, the choice of random seed is more meaningful when $k=5$ than when $k=200$, in the context of the specific choices of $R$, $S$, $d$, $\delta$, and all other modeling decisions that are common across $\Theta$. Importantly, this is a claim that does not require us to understand the choice of random seed within the narrow constraints of particular performance metrics, the selection of which would constitute additional modeling decisions to be accounted for. 

On the one hand, we might use these results to guide our selection of $k$, helping us to narrow down the selection of models from the full set of $\Theta$ to a more manageable subset or even a single model. However, we could also consider the extent to which structural variation presents greater information about $R[S]$ across the particular interpretive lenses of each $\theta$ and so working with a larger sample of the structural space. 

\subsection{The meaning of \emph{k}}

We now consider the meaning of the choice of $k$ by exploring the magnitude of structural differences that result from varying this choice. To some, the choice of $k$ may be thought of as highly significant, and they are therefore likely to use evaluation measures as evidence for one choice over another. However, there is also a view that changing $k$ may change the latent attributes in a way that is more cosmetic than semantic, and so may constitute an interpretively neutral decision. For example, it has been suggested that the choice of $k$ may be seen as a choice of topic specificity \citep{10.3389/frai.2020.00062}. Here, we sketch out a starting point for analyzing the structural consequences of choosing $k$ within the context of our chosen dataset. 

From the above analysis, we have $f(S \mid \theta R, d)$ for each $\theta \in \Theta$. We abbreviate the set of structures that share the same choice of $k$ as follows:

\begin{equation}
    \mathcal{F}_k = \big\{f(S \mid \theta R,\, d)  
    \, : \,
    \theta \in \Theta_{k}
    \big\}.
\end{equation}

For example $\mathcal{F}_{k=5}$ is the set of JSD matrices obtained from $\Theta_{k=5}$, and we can treat it as a sample of all LDA models with $k=5$ applied to the dataset.  

We let $\bar{\delta}(\mathcal{F}_{k_i}, \mathcal{F}_{k_j})$ be the average of the Procrustes distances between the elements of $\mathcal{F}_{k_i}$ and $\mathcal{F}_{k_j}$. For example, $\bar{\delta}(\mathcal{F}_{k=5}, \mathcal{F}_{k=15})$ would be the average Procrustes distance between all pairs of $f(S \mid \theta_{k=5}R, d)$ and $f(S \mid \theta_{k=15}R, d)$. To contextualize these average structural distances, we also include the average self Procrustes distance of each group, with $\bar{\delta}(\mathcal{F}_{k_i}, \mathcal{F}_{k_i})$ being the average of $f(\Psi_{k_i} \mid \Phi, \delta)$ as defined in the previous illustration. 

\begin{figure}
  \centering
  \includegraphics[width=\linewidth]{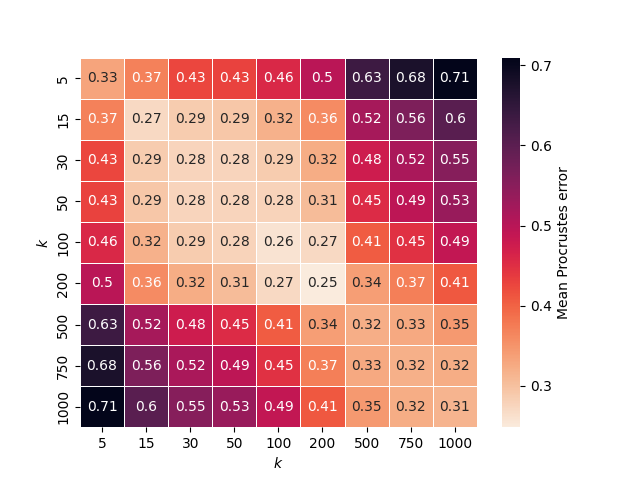}
  \caption{A heatmap of the mean Procrustes error between LDA models across different dimensionalities, $k$, with the diagonal showing the mean Procrustes distance between models with the same value of $k$.}
  \label{fig:annotated_heatmap}
\end{figure}

The results for all pairs of $k$ are given in figure \ref{fig:annotated_heatmap}. We can see that the average difference between $k=100$ and $k=200$ is relatively small, with $\bar{\delta}(\mathcal{F}_{k=100}, \mathcal{F}_{k=200})$ being 0.27 compared to 0.26 for $\bar{\delta}(\mathcal{F}_{k=100}, \mathcal{F}_{k=100})$ and 0.25 for $\bar{\delta}(\mathcal{F}_{k=200}, \mathcal{F}_{k=200})$. 

From here, additional analyses might be done to more deeply understand the semantics of $k$ in this context. For example, these structures might all be clustered in some way to give a general sense of where the general structure of $S$ most changes. For example, it is possible that $k =$ 15, 30, and 50 might be usefully seen as one structural flavor and that $k =$ 500, 750, and 1,000 constitute a distinct flavor. 

\subsection{The relationship between Procrustes distance and Pearson correlation}

Here, we consider the relationship between measuring structural relationships using the Procrustes distance, which we denote by $\delta_{\mathrm{proc}}$, and the using the Pearson correlation coefficient, denoted by $\delta_{\mathrm{pear}}$. Recall that we have 90 structures on $S$, corresponding to each of the 90 models in $\Theta$ and that we have calculated Procrustes distance between all model pairs. For the same model pairs, we calculate the Pearson correlation coefficient.  If the Procrustes distance and Pearson correlation both reflect the structural relationships between two dissimilarity matrices, then we would expect agreement between them to look like negative correlation, given that the Pearson correlation measures similarity and the Procrustes distance measures dissimilarity. The larger the differences we observe between models according to $\delta_{\mathrm{proc}}$, then the smaller their similarity ought to be according to $\delta_\mathrm{pear}$, and vice versa. 

\begin{figure}
  \centering
  \includegraphics[width=\linewidth]{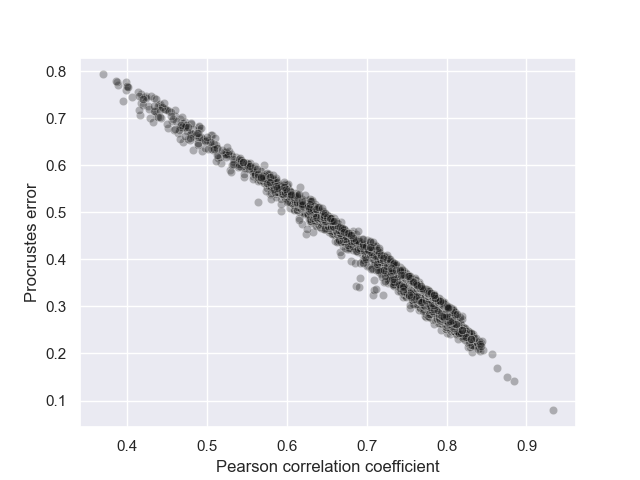}
  \caption{For each pair of LDA models compared, their Procrustes error (vertical axis) and Pearson correlation coefficient (horizontal axis). The Pearson correlation coefficient between them is -0.99 for 4,005 model pairs.}
  \label{fig:pearson_procrustes_scatterplot}
\end{figure}

We find that the two measures are indeed negatively correlated, having a Pearson correlation coefficient of -0.99 (see figure \ref{fig:pearson_procrustes_scatterplot}), suggesting that using one over the other is unlikely to change the structure of $f(\Theta \mid \Phi, \delta)$. Note that this numeric summarization of the relationship between $\delta_{\mathrm{proc}}$
and $\delta_{\mathrm{pear}}$ depends, at least some extent, on the choice of the Pearson correlation to be $\delta'$, the outer structural relationship measure. This is a useful reminder that there is always some unaccounted for modeling decision such that, even for computers, there is no perspective that is perfectly neutral. If we were to continue this analysis by extending it to a larger set of structural relation measures, the semantics of $\delta$ that we would find would still depend on $\delta'$ to an unknown degree. However, this framework provides a starting point from which we might begin making rigorous claims about the semantic consequences of our modeling decisions. 

\section{Discussion}
\label{sec:Discussion}

We now consider a few implications of this framework and some large-scale questions that we might collectively begin answering. 

\subsection{Navigating modeling decisions}

In the context of this theoretical framework, model evaluation has the purpose of constraining the space of modeling decisions that a researcher makes so as to increase the usefulness of the resulting representation or structure. However, defining what is useful is nontrivial. 

Further, semantic modeling is inherently an unsupervised problem in this framework (since the target variable is defined to be sign relations). Here, performance measures and benchmarks do not tell us about the optimal or uniquely correct interpretations of signs, but rather allow us to describe interpretations with respect to the well-defined ends of those measures. Then performance measures can be understood as a kind of selection pressure exerted on the space of potential structures. 

An important question is how semantically distinct different performance measures are, which we can describe as the distinctness of the idealized structures they entail. If a performance measure, $p_i$, entails an ideal structure (or set of structures), $f^*(S \mid p_i)$, then its relationship to another measure $p_j$ is given by the regularities that emerge between $f^*(S \mid p_i)$ and $f^*(S \mid p_j)$ across various contexts. 

For example, if large language models tend to utilize the same set of performance measures, then it is possible that their interpretive dispositions may come to resemble each other even if they differ with respect to other modeling decisions. 

\subsection{Cultural models of mathematics}
Our primary concern in proposing this framework has been the making of rigorous, clear claims about models in a relational, symbolic fashion. While we have not been concerned with measures of performance, we have also not been concerned with the direct interpretation of sign structures. If we think of the computational analysis of complex semiotic systems as requiring a translation from the cultural domain to the computational domain \citep{stine2025semioticcomplexity}, then we have yet to consider how sign structures may be translated back into the cultural. Doing so would require reading sign geometries as a novel kind of text. While we have not considered how to do such a reading in this paper, it is useful to consider the sorts of questions we might answer by doing so. 

In the previous section, we suggested that this framework might be used to investigate the structural dispositions of performance measures. Suppose one could usefully read structures on $S$. Then one could build up a picture of the way one performance measure tends to structure signs on the basis of how one reads the structures it produces. Here, the modeling direction is reversed---rather than model culture with mathematics, this would be closer to modeling mathematics with culture. 

As another example, one might build up a cultural picture of a large language model by reading the structures it imposes on various representations and identifying the interpretive regularities that distinguish it from other kinds of models. This would enable another dimension in which we understand models beyond their mechanics. Further, it would make visible the values underlying large language models, and by extension the companies that create and deploy them, since their interpretive dispositions encode their creators' ends (see section \ref{subsec:symbols_and_ends}). This is especially important if such models become part of how we read latent structures back into the cultural domain, as suggested by \cite{kommers2025meaningmetricusingllms}.

\section{Conclusion}

\cite{schmidt2016digital} asks an important question about the extent to which computational humanists ought to understand the inner workings of their models and algorithms. This is an important question, since our analyses do not tell us only about our data, but also about our methods. While we will likely continue to see a proliferation of highly complicated model architectures emerge, the comparative framework we have proposed should allow any researcher to assess the functional consequences of choosing one model over some other, regardless of one's technical expertise.

If we as computational modelers of human meaning-making see usefulness in the distant reading of our data through models, then it is important that we consider the extent to which we are always reading an inseparable combination of data and model. Therefore, an important consideration for this field is the extent to which any results we obtain from an analysis are a reflection of our data versus the models through which we ``read" those data. By proposing this framework, we hope to contribute a starting point for disentangling this relationship through a distant reading of the modeling decisions we make as a new kind of computational hermeneutics.

\section*{Acknowledgments}
This material is based upon work supported by the National Science Foundation under Grant No. OIA-1946391. Any opinions, findings, and conclusions or recommendations expressed in this material are those of the authors and do not necessarily reflect the views of the National Science
Foundation.

\bibliographystyle{plainnat}
\bibliography{bibliography}

\end{document}